\documentclass[final]{cvpr}

\usepackage{times}
\usepackage{epsfig}
\usepackage{graphicx}
\usepackage{amsmath}
\usepackage{amssymb}
\usepackage{bbding} 
\usepackage{subfigure}
\usepackage{bm}
\usepackage{enumerate}
\usepackage{multirow}
\usepackage{multicol}
\usepackage{colortbl}
\usepackage{color}
\usepackage{mathtools}
\usepackage{booktabs}
\usepackage{cuted}
\usepackage{capt-of}

\usepackage[pagebackref=true,breaklinks=true,colorlinks,bookmarks=false]{hyperref}

\def\cvprPaperID{4405} 
\def\confYear{CVPR 2021}
\begin{document}
	
	\title{UPFlow: Upsampling Pyramid for Unsupervised Optical Flow Learning}
	
	\author{%
		Kunming Luo$^1$ \quad
		Chuan Wang$^1$ \quad
		Shuaicheng Liu$^{2,1}$\thanks{Corresponding author}  \quad
		Haoqiang Fan$^1$\quad
		Jue Wang$^1$ \quad
		Jian Sun$^1$ \\\\
		$^1$Megvii Technology \quad
		$^2$University of Electronic Science and Technology of China
		\\
		\url{https://github.com/coolbeam/UPFlow_pytorch}
	}

	\maketitle

	\begin{abstract}\label{sec:abs}
		We present an unsupervised learning approach for optical flow estimation by improving the upsampling and learning of pyramid network. 
		We design a self-guided upsample module to tackle the interpolation blur problem caused by bilinear upsampling between pyramid levels. Moreover, we propose a pyramid distillation loss to add supervision for intermediate levels via distilling the finest flow as pseudo labels.
		By integrating these two components together, our method achieves the best performance for unsupervised optical flow learning on multiple leading benchmarks, including MPI-SIntel, KITTI 2012 and KITTI 2015. 
		In particular, we achieve EPE=1.4 on KITTI 2012 and F1=9.38\% on KITTI 2015, which outperform the previous state-of-the-art methods by 22.2\% and 15.7\%, respectively. 
	\end{abstract} 
	\section{Introduction}\label{sec:intro}
\begin{figure}[t]
		\centering
		\includegraphics[width=1.0\linewidth]{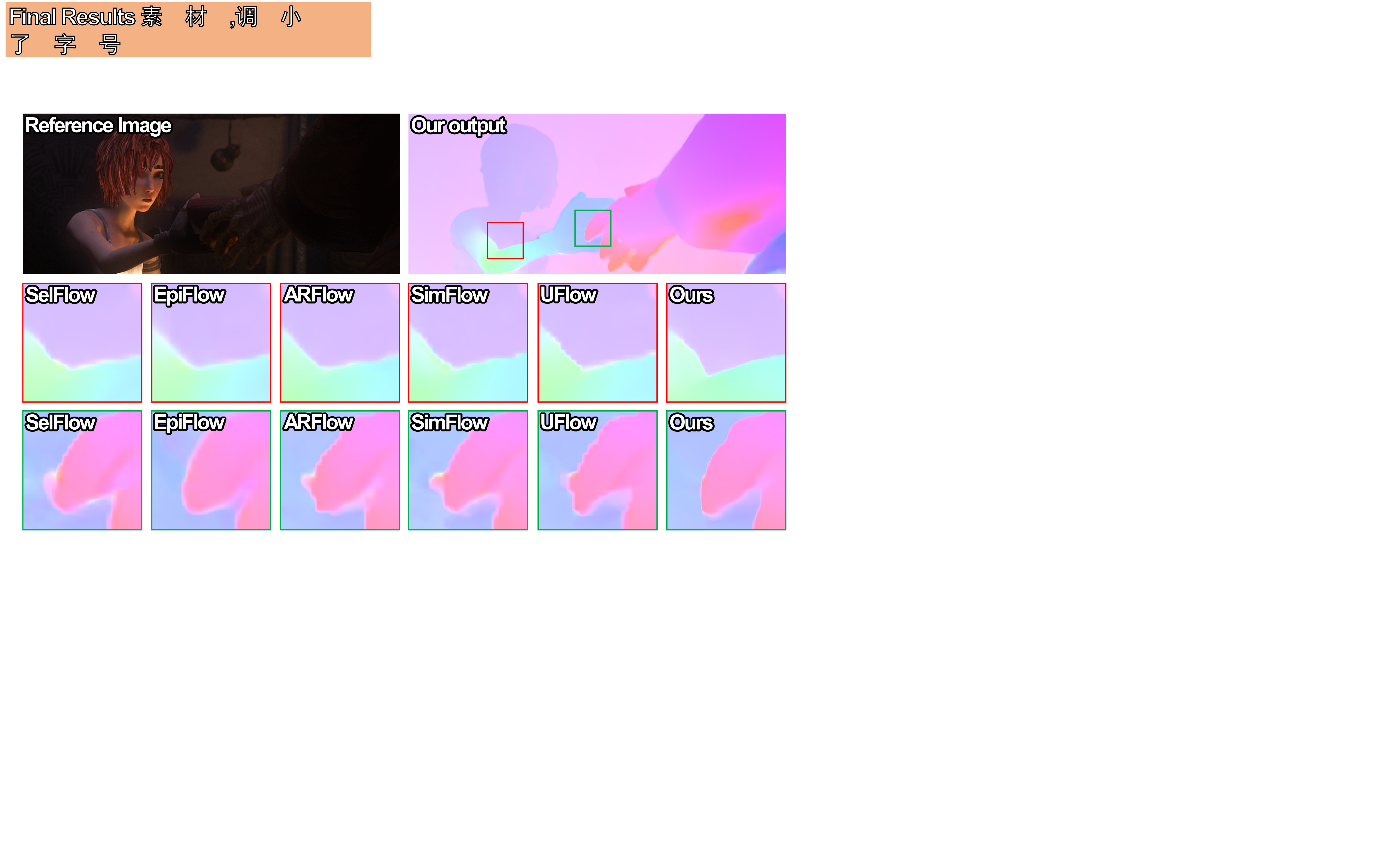}
		\caption{An example from Sintel Final benchmark. Compared with previous unsupervised methods including SelFlow~\cite{Liu2019CVPR}, EpiFlow~\cite{Epipolar_flow_2019cvpr}, ARFlow~\cite{liu2020learning}, SimFlow~\cite{simFlow2020eccv} and UFlow~\cite{jonschkowski2020matters}, our approach produces sharper and more accurate results in object edges. 
		}\label{fig:teaser_sintel_final}
	\end{figure}

	Optical flow estimation has been a fundamental computer vision task for decades, which has been widely used in various applications such as video editing~\cite{Jiang_2018_CVPR}, behavior recognition~\cite{Simonyan2014} and object tracking~\cite{largemotion_network_design_iccv2017}. 
	The early solutions focus on minimizing a pre-defined energy function with optimization tools~\cite{Thomas2004,Sun2010,EpicFlow_2015}. Nowadays deep learning based approaches become popular, which can be classified into two categories, the supervised~\cite{FlowNet2,spynet2017} and unsupervised ones~\cite{Ren2017aaai,wang2018}. 
	The former one uses synthetic or human-labelled dense optical flow as ground-truth to guide the motion regression. 
	The supervised methods have achieved leading performance on the benchmark evaluations. However, the acquisition of ground-truth labels are expensive. In addition, the generalization is another challenge when trained on synthetic datasets. 
	As a result, the latter category, \ie the unsupervised approaches attracts more attentions recently, which does not require the ground-truth labels. 
	In unsupervised methods, the photometric loss between two images is commonly used to train the optical flow estimation network. To facilitate the training, the pyramid network structure~\cite{pwc_net,irrpwc} is often adopted, such that both global and local motions can be captured in a coarse-to-fine manner. 
	However, there are two main issues with respect to the pyramid learning, which are often ignored previously. We refer the two issues as \textbf{bottom-up} and \textbf{top-down} problems.
	The bottom-up problem refers to the upsampling module in the pyramid.
	Existing methods often adopt simple bilinear or bicubic upsampling~\cite{liu2020learning,jonschkowski2020matters}, which interpolates cross edges, resulting in blur artifacts in the predicted optical flow. Such errors will be propagated and aggregated when the scale becomes finer. 
	Fig.~\ref{fig:teaser_sintel_final} shows an example. 
	The top-down problem refers to the pyramid supervision. The previous leading unsupervised methods typically add guidance losses only on the final output of the network, while the intermediate pyramid levels have no guidance. In this condition, the estimation errors in coarser levels will accumulate and damage the estimation at finer levels due to the lack of training guidance. 
	
	
	To this end, we propose an enhanced pyramid learning framework of unsupervised optical flow estimation. First, we introduce a self-guided upsampling module that supports blur-free optical flow upsampling by using a self-learned interpolation flow instead of the straightforward interpolations. Second, we design a new loss named pyramid distillation loss that supports explicitly learning of the intermediate pyramid levels by taking the finest output flow as pseudo labels.
	To sum up, our main contributions include:
	\begin{itemize}
		\item We propose a \textbf{self-guided upsampling module} to tackle the interpolation problem in the pyramid network, which can generate the sharp motion edges.
		\item We propose a \textbf{pyramid distillation loss} to enable robust supervision for unsupervised learning of coarse pyramid levels.
		\item We achieve superior performance over the state-of-the-art unsupervised methods with a relatively large margin, validated on multiple leading benchmarks. 
		
	\end{itemize}
	
\section{Related Work}\label{sec:related_work}
\subsection{Supervised Deep Optical Flow}\label{sec:related_supervised_methods}
Supervised methods require annotated flow ground-truth to train the network~\cite{cnn_patch_match2017,zhao2020maskflownet,CostVolume_cvpr2017,VolumetricCN2019}.  FlowNet~\cite{Flownet_flyingchairs} was the first work that proposed to learn optical flow estimation by training fully convolutional networks on synthetic dataset FlyingChairs. 
Then, FlowNet2~\cite{FlowNet2} proposed to iteratively stack multiple networks for the improvement.
To cover the challenging scene with large displacements, SpyNet~\cite{spynet2017} built a spatial pyramid network to estimate optical flow in a coarse-to-fine manner. 
PWC-Net~\cite{pwc_net} and LiteFlowNet~\cite{LiteFlowNet} proposed to build efficient and lightweight networks by warping feature and calculating cost volume at each pyramid level. 
IRR-PWC~\cite{irrpwc} proposed to design pyramid network by an iterative residual refinement scheme.
Recently, RAFT~\cite{raft2020} proposed to estimate flow fileds by 4D correlation volume and recurrent network, yielding state-of-the-art performance.
In this paper, we work in unsupervised setting where no ground-truth labels are required.

\subsection{Unsupervised Deep Optical Flow} \label{sec:related_unsup_methods}
Unsupervised methods do not need annotations for training~\cite{unsup_ICIP_2016,Jason2016}, which can be divided into two categories: the occlusion handling methods and the alignment learning methods. The occlusion handling methods mainly focus on excluding the impact of the occlusion regions that cannot be aligned inherently. For this purpose, many methods are proposed, including the occlusion-aware losses by forward-backward occlusion checking~\cite{unflow_2018aaai} and range-map occlusion checking~\cite{wang2018}, the data distillation methods~\cite{Pengpeng2019,Liu2019CVPR,tip_STFlow}, and augmentation regularization loss~\cite{liu2020learning}. On the other hand, the alignment learning methods are mainly developed to improve optical flow learning under multiple image alignment constrains, including the census transform constrain~\cite{Ren2017aaai}, multi-frame formulation~\cite{unflow_multi_occ}, epipolar constrain~\cite{Epipolar_flow_2019cvpr}, depth  constrains~\cite{Anurag2019,dfnet_zou_2018,Yin2018GeoNetUL,Liu2019UnsupervisedLO} and feature similarity constrain~\cite{simFlow2020eccv}. Recently, UFlow~\cite{jonschkowski2020matters} achieved the state-of-the-art performance on multiple benchmarks by systematically analyzing and integrating multiple unsupervised components into a unified framework. In this paper, we propose to improve optical flow learning with our improved pyramid structure. 

\subsection{Image Guided Optical Flow Upsampling} \label{sec:related_image_guided_upsample}
A series of methods have been developed to upsample images, depths or optical flows by using the guidance information extracted from high resolution images. The early works such as joint bilateral upsampling~\cite{kopf2007JBU} and guided image filtering~\cite{he2010guided} proposed to produce upsampled results by filters extracted from the guidance images. Recent works~\cite{li2016fastfgi,wu2018fast,su2019pac} proposed to use deep trainable CNNs to extract guidance feature or guidance filter for upsampling. In this paper, we build an efficient and lightweight self-guided upsampling module to extract interpolation flow and interpolation mask for optical flow upsampling. By inserting this module into a deep pyramid network, high quality results can be obtained. 

\begin{figure*}
	\centering
	\includegraphics[width=1.0\linewidth]{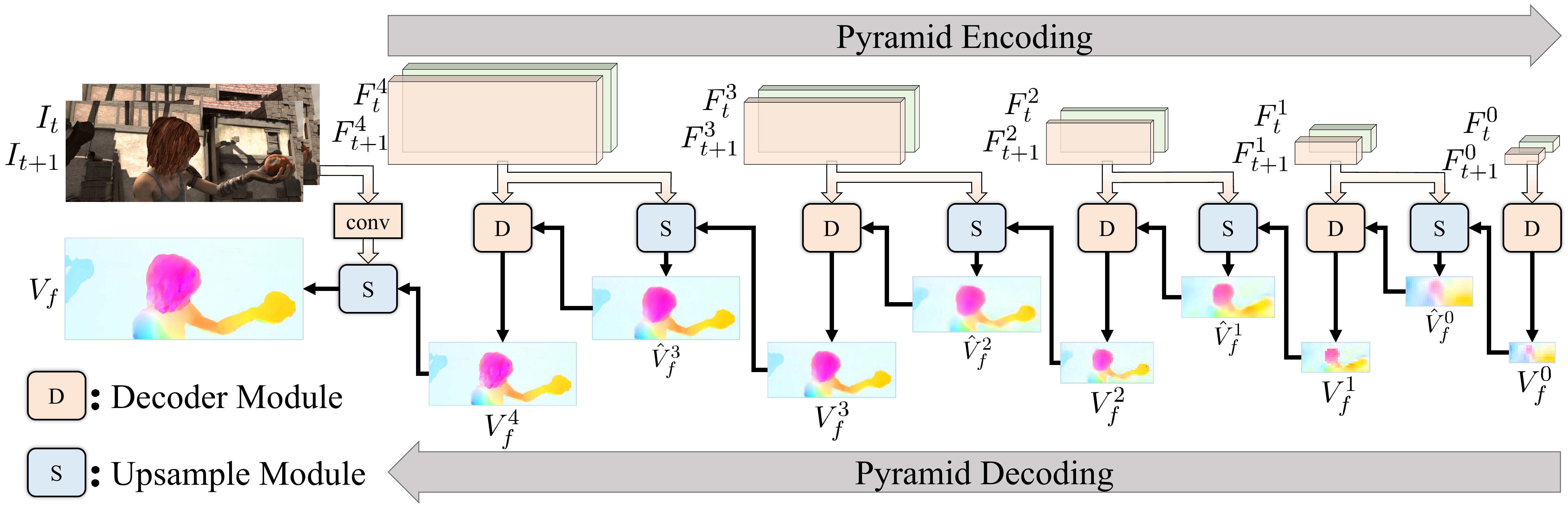}
	\caption{Illustration of the pipeline of our network, which contains two stage: pyramid encoding to extract feature pairs in different scales and pyramid decoding to estimate optical flow in each scale. Note that the parameters of the decoder module and the upsample module are shared across all the pyramid levels. }\label{fig:algo_our_pipeline}
\end{figure*}

\section{Algorithm}\label{sec:algo}
\subsection{Pyramid Structure in Optical Flow Estimation}\label{sec:algo_pyramid_structure}
Optical flow estimation can be formulated as: 
\begin{equation}\label{eq:optical_flow_model}
	V_{f}=\mathcal{H}(\theta,I_{t},I_{t+1}), 
\end{equation}
where $I_{t}$ and $I_{t+1}$ denote the input images, $\mathcal{H}$ is the estimation model with parameter $\theta$, and $V_{f}$ is the forward flow field that represents the movement of each pixel in $I_t$ towards its corresponding pixel in $I_{t+1}$. 

The flow estimation model $\mathcal{H}$ is commonly designed as a pyramid structure, such as the classical PWC-Net~\cite{pwc_net}. The pipeline of our network is illustrated in Fig.~\ref{fig:algo_our_pipeline}. 
The network can be divided into two stages: pyramid encoding and pyramid decoding. 
In the first stage, we extract feature pairs in different scales from the input images by convolutional layers. 
In the second stage, we use a decoder module $\mathcal{D}$ and an upsample module $\mathcal{S}_{\uparrow}$ to estimate optical flows in a coarse-to-fine manner. 
The structure of the decoder module $\mathcal{D}$ is the same as in UFlow~\cite{jonschkowski2020matters}, which contains feature warping, cost volume construction by correlation layer, cost volume normalization, and flow decoding by fully convolutional layers. Similar to recent works~\cite{irrpwc,jonschkowski2020matters}, we also make the parameters of $\mathcal{D}$ and $\mathcal{S}_{\uparrow}$ shared across all the pyramid levels. In summary, the pyramid decoding stage can be formulated as follows: 
\begin{align}
	\hat{V}_{f}^{i-1} & =  \mathcal{S}_{\uparrow}(F_{t}^{i},F_{t+1}^{i},V_{f}^{i-1}),  \label{eq:pyramid_structure_upsample} \\
	V_{f}^{i} & =\mathcal{D}(F_{t}^{i},F_{t+1}^{i},\hat{V}_{f}^{i-1}),  \label{eq:pyramid_structure_decode}
\end{align}
where $i\in \{0,1,...,N\}$ is the index of each pyramid level and the smaller number represents the coarser level, $F_{t}^{i}$ and $F_{t+1}^{i}$ are features extracted from $I_t$ and $I_{t+1}$ at the $i$-th level, and $\hat{V}_{f}^{i-1}$ is the upsampled flow of the $i-1$ level. 
In practice, considering the accuracy and efficiency, $N$ is usually set to 4~\cite{irrpwc,pwc_net}. 
The final optical flow result is obtained by directly upsampling the output of the last pyramid level. 
Particularly, in Eq.~\ref{eq:pyramid_structure_upsample}, the bilinear interpolation is commonly used to upsample flow fields in previous methods~\cite{irrpwc,jonschkowski2020matters}, which may yield noisy or ambiguity results at object boundaries. In this paper, we present a self-guided upsample module to tackle this problem as detailed in Sec.~\ref{sec:algo_self_guided_upsample}. 

\begin{figure}
	\centering
	\includegraphics[width=0.45\textwidth]{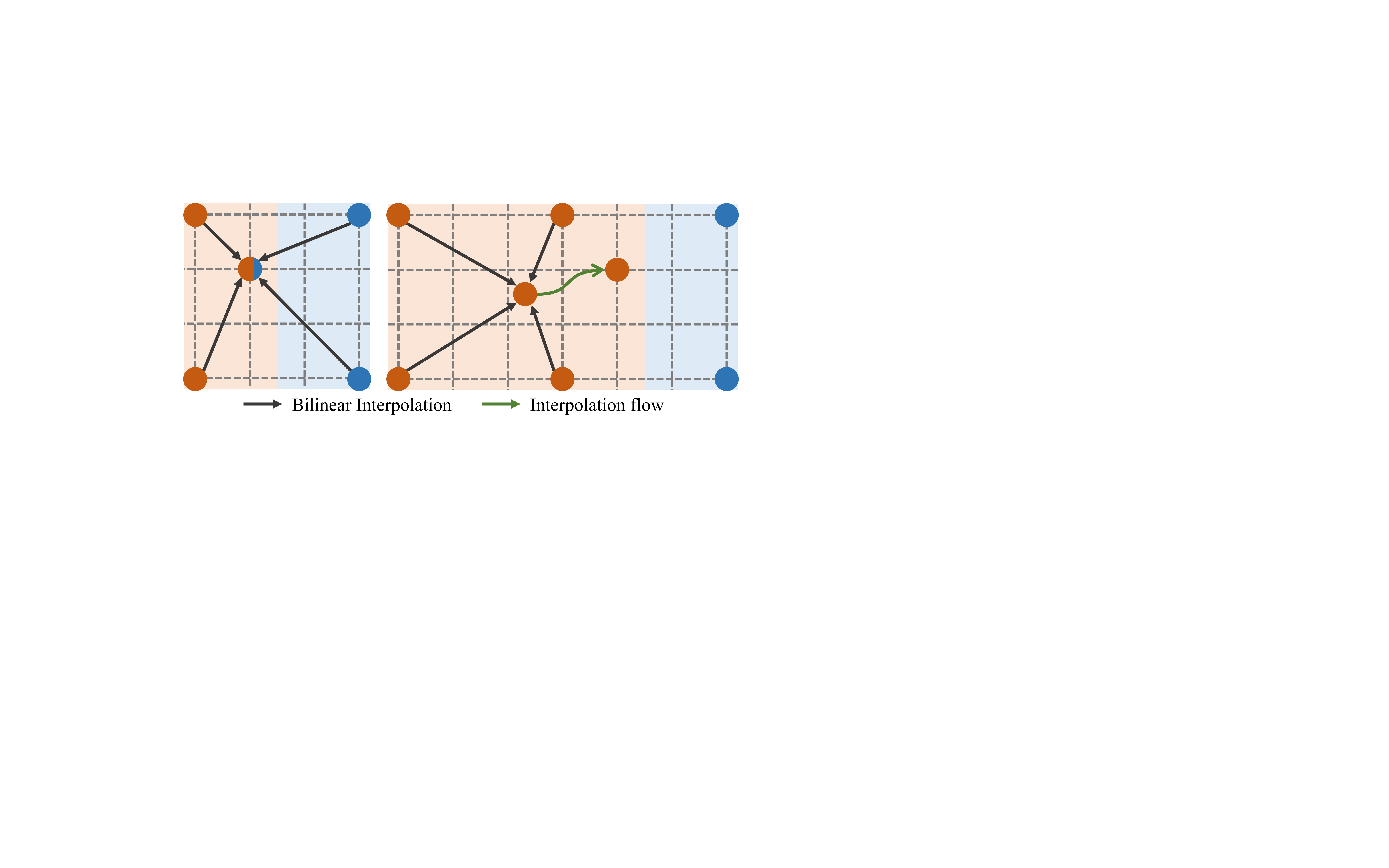}
	\caption{
		Illustration of bilinear upsampling (left) and the idea of our self-guided upsampling (right). Red and blue dots are motion vectors from different objects.
		Bilinear upsampling often produces cross-edge interpolation. We propose to first interpolate a flow vector in other position without crossing edge and then bring it to the desired position by our learned interpolation flow. 
	}\label{fig:algo_self_guided_interpolation_}
\end{figure}

\subsection{Self-guided Upsample Module}\label{sec:algo_self_guided_upsample}
In Fig.~\ref{fig:algo_self_guided_interpolation_} left, the case of bilinear interpolation is illustrated. We show $4$ dots which represent $4$ flow vectors, belonging to two motion sources, marked as red and blue, respectively. The missing regions are then bilinear interpolated with no semantic guidance. Thus, a mixed interpolation result is generated at the red motion area, resulting in cross-edge interpolation. 
In order to alleviate this problem, we propose a self-guided upsample module (SGU) to change the interpolation source points by an interpolation flow.
The main idea of our SGU is shown in Fig.~\ref{fig:algo_self_guided_interpolation_} right. We first interpolate a point by its enclosing red motions and then bring the result to the target place with the learned interpolation flow (Fig.~\ref{fig:algo_self_guided_interpolation_}, green arrow). As a result, the mixed interpolation problem can be avoided. 

\begin{figure}
	\centering
	\includegraphics[width=0.45\textwidth]{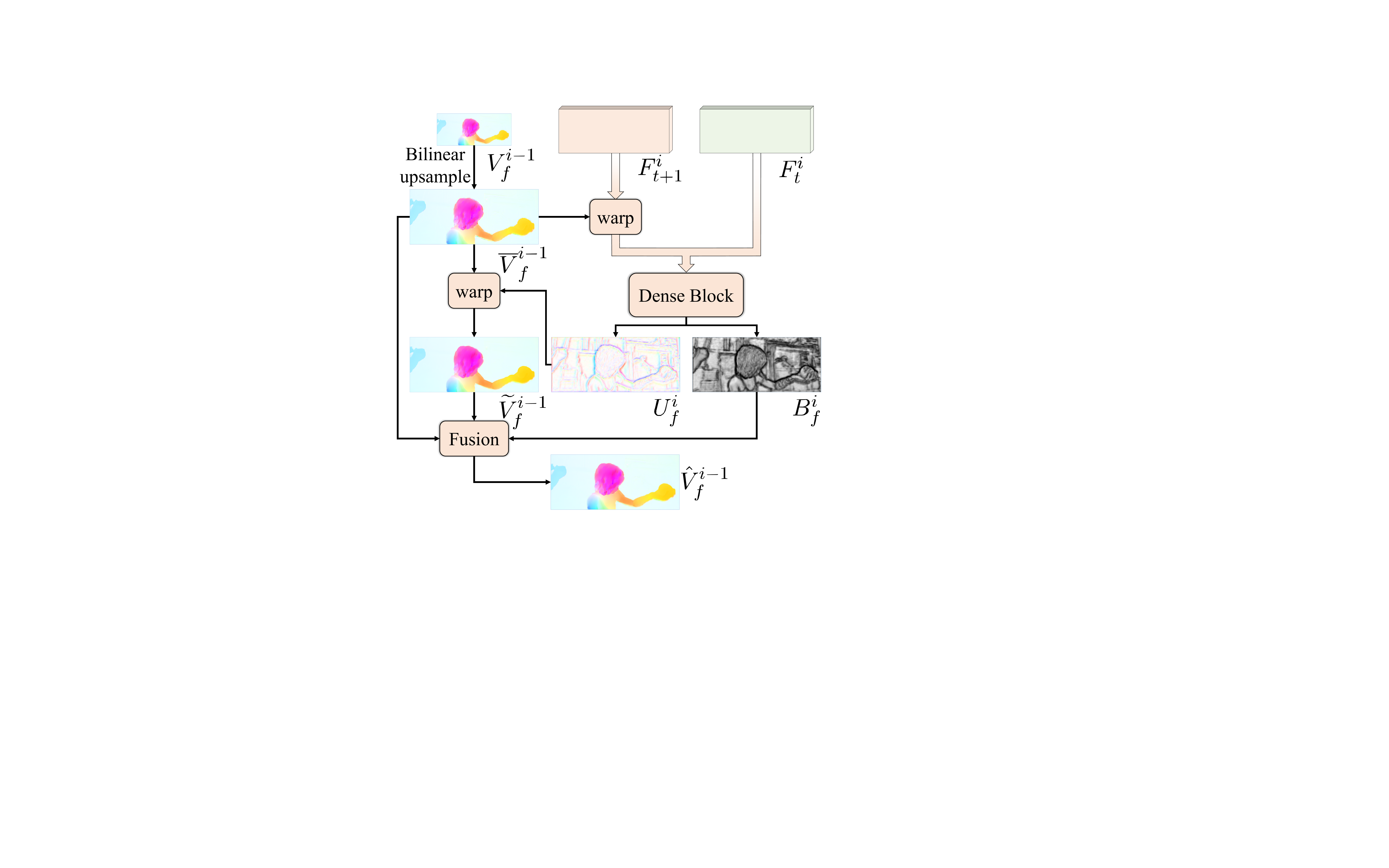}
	\caption{Illustration of our self-guided upsample module. We first upscale the input low resolution flow $V_{f}^{i-1}$ by bilinear upsampling and use a dense block to compute an interpolation flow $U_{f}^{i}$ and an interpolation map $B_{f}^{i}$. Then we generate the high resolution flow by warping and fusion. 
	}\label{fig:algo_our_upsample_model}
\end{figure}

\begin{figure*}
	\centering
	\includegraphics[width=0.98\textwidth]{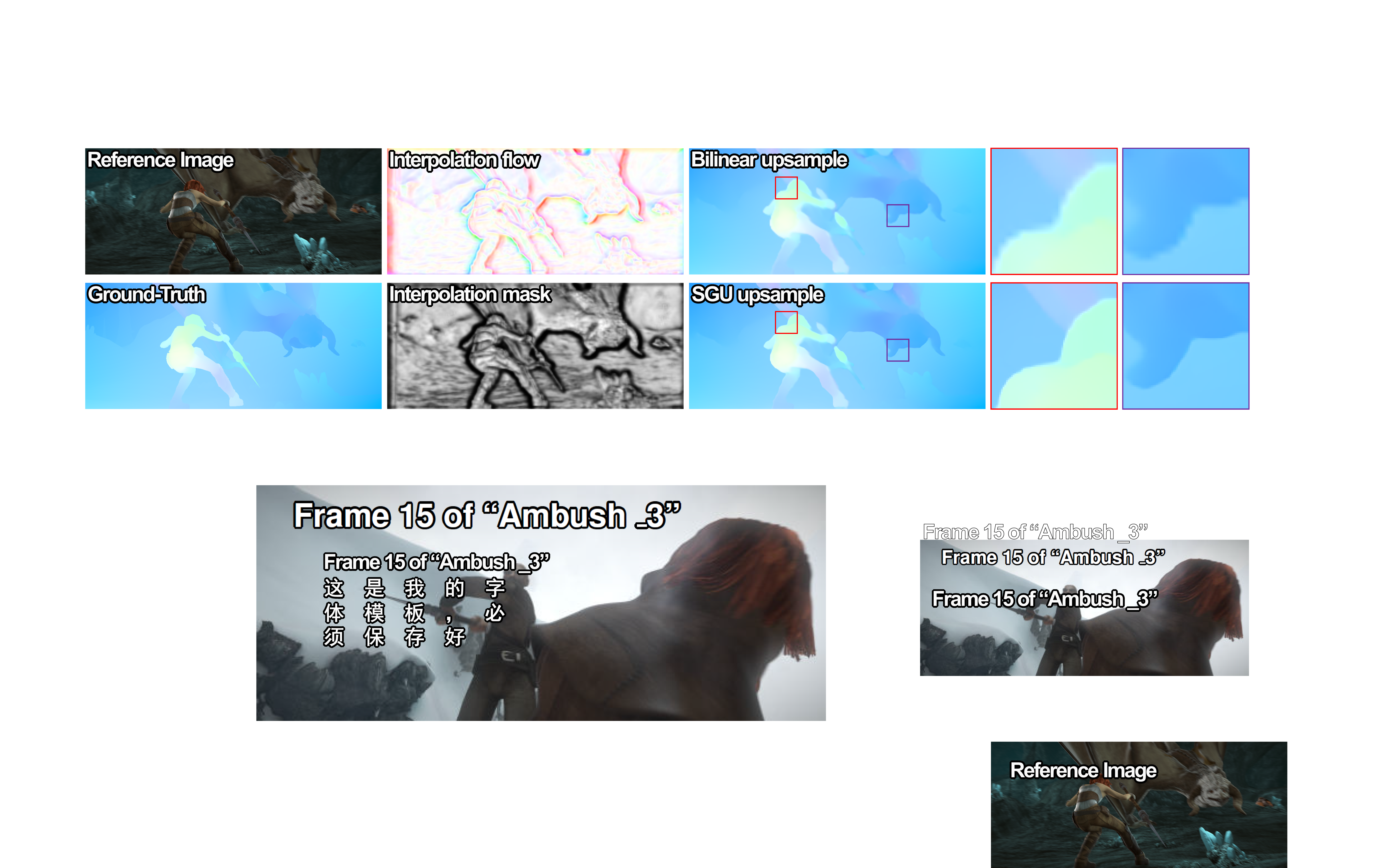}
	\caption{Visual example of our self-guided upsample module (SGU) on MPI-Sintel Final dataset. Results of bilinear method and our SGU are shown. The zoom-in patches are also shown on the right of each sample for better comparison. 
	}\label{fig:upsample_inter_res_sintel}
\end{figure*}

In our design, to keep the interpolation in plat areas from being changed and make the interpolation flow only applied on motion boundary areas, we learn a per-pixel weight map to indicate where the interpolation flow should be disabled. Thus, the upsampling process of our SGU is a weighted combination of the bilinear upsampled flow and a modified flow obtained by warping the upsampled flow with the interpolation flow. 
The detailed structure of our SGU module is shown in Fig.~\ref{fig:algo_our_upsample_model}. 
Given a low resolution flow $V_{f}^{i-1}$ from the $i-1$-th level, we first generate an initial flow $\overline{V}_{f}^{i-1}$ in higher resolution by bilinear interpolation:
\begin{equation}\label{eq:bilinear_interpolation_upsample}
	\overline{V}_{f}^{i-1}(\bm{p})=\sum_{\bm{k} \in \mathcal{N}(\bm{p}/s)} w(\bm{p}/s, \bm{k}) V_{f}^{i-1}(\bm{k}), 
\end{equation}
where $\bm{p}$ is a pixel coordinate in higher resolution, $s$ is the scale magnification, $\mathcal{N}$ denotes the four neighbour pixels, and $w(\bm{p}/s, \bm{k})$ is the bilinear interpolation weights. 
Then, we compute an interpolation flow $U_{f}^{i}$ from features $F_t^{i}$ and $F_{t+1}^{i}$ to change the interpolation of $\overline{V}_{f}^{i-1}$ by warping: 
\begin{align}
	\widetilde{V}_{f}^{i-1}(\bm{p}) &= \sum_{\bm{k} \in \mathcal{N}(\bm{d})} w(\bm{d},\bm{k} ) \overline{V}_{f}^{i-1}(\bm{k}), \label{eq:guided_interpolation_upsample} \\
	\bm{d} &= \bm{p}+U_{f}^{i}(\bm{p}), \label{eq:guided_interpolation_upsample_}
\end{align}
where $\widetilde{V}_{f}^{i-1}$ is the result of warping $\overline{V}_{f}^{i-1}$ by the interpolation flow $U_{f}^{i}$. Since the interpolation blur only occurs in object edge regions, it is unnecessary to learn interpolation flow in flat regions. We thus use an interpolation map $B_{f}^{i}$ to explicitly force the model to learn interpolation flow only in motion boundary regions. 
The final upsample result is the fusion of $\widetilde{V}_{f}^{i-1}$ and $\overline{V}_{f}^{i-1}$:
\begin{equation}\label{eq:guided_interpolation_upsample_fusion}
	\hat{V}_{f}^{i-1}=B_{f}^{i}\odot \overline{V}_{f}^{i-1}+(1-B_{f}^{i})\odot \widetilde{V}_{f}^{i}, 
\end{equation}
where $\hat{V}_{f}^{i-1}$ is the output of our self-guided upsample module and $\odot$ is the element-wise multiplier. 

To produce the interpolation flow $U_{f}^{i}$ and the interpolation map $B_{f}^{i}$, we use a dense block with $5$ convolutional layers. Specifically, we concatenate the feature map $F_{t}^{i}$ and the warped feature map $F_{t+1}^{i}$ as the input of the dense block. The kernel number of each convolutional layer in the dense block is $32$, $32$, $32$, $16$, $8$ respectively. The output of the dense block is a tensor map with \textbf{3 channels}. We use the first two channels of the tensor map as the interpolation flow and use the last channel to form the interpolation map through a sigmoid layer. 
Note that, no supervision is introduced for the learning of interpolation flow and interpolation map. Fig.~\ref{fig:upsample_inter_res_sintel} shows an example from MPI-Sintel Final dataset, where our SGU produces cleaner and sharper results at object boundaries compared with the bilinear method. Interestingly, the self-learned interpolation map is nearly to be an edge map and the interpolation flow is also focused on object edge regions. 
\begin{figure*}[ht]
	\centering
	\subfigure[Visual comparison on KITTI 2012 (first two rows) and KITTI 2015 (last two rows).]{
		\includegraphics[width=0.98\linewidth]{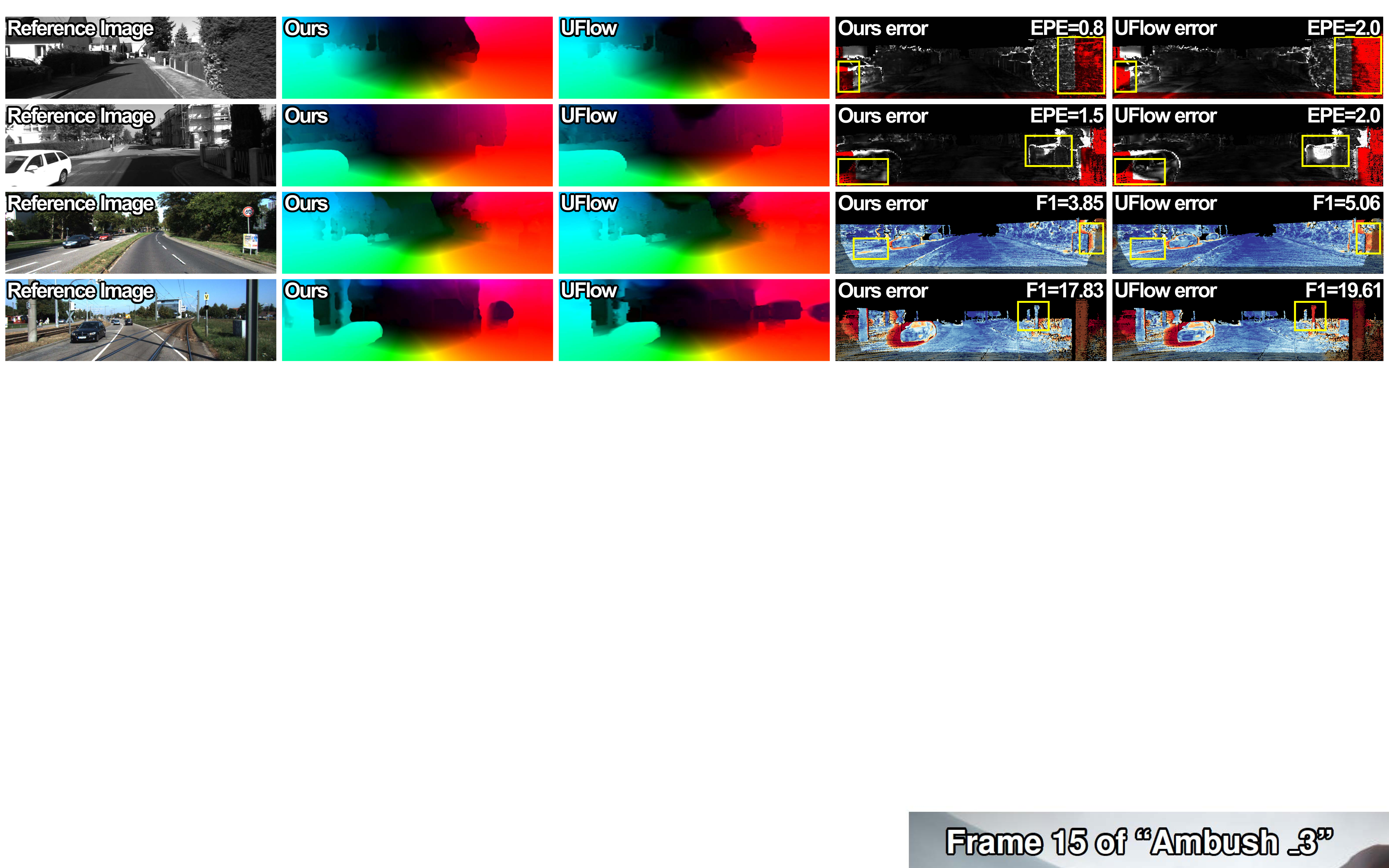}
		\label{fig:results_pk_benchmark_kitti}
	}
	\subfigure[Visual comparison on Sintel Clean (first two rows) and Sintel Final (last two rows).]{
		\includegraphics[width=0.98\linewidth]{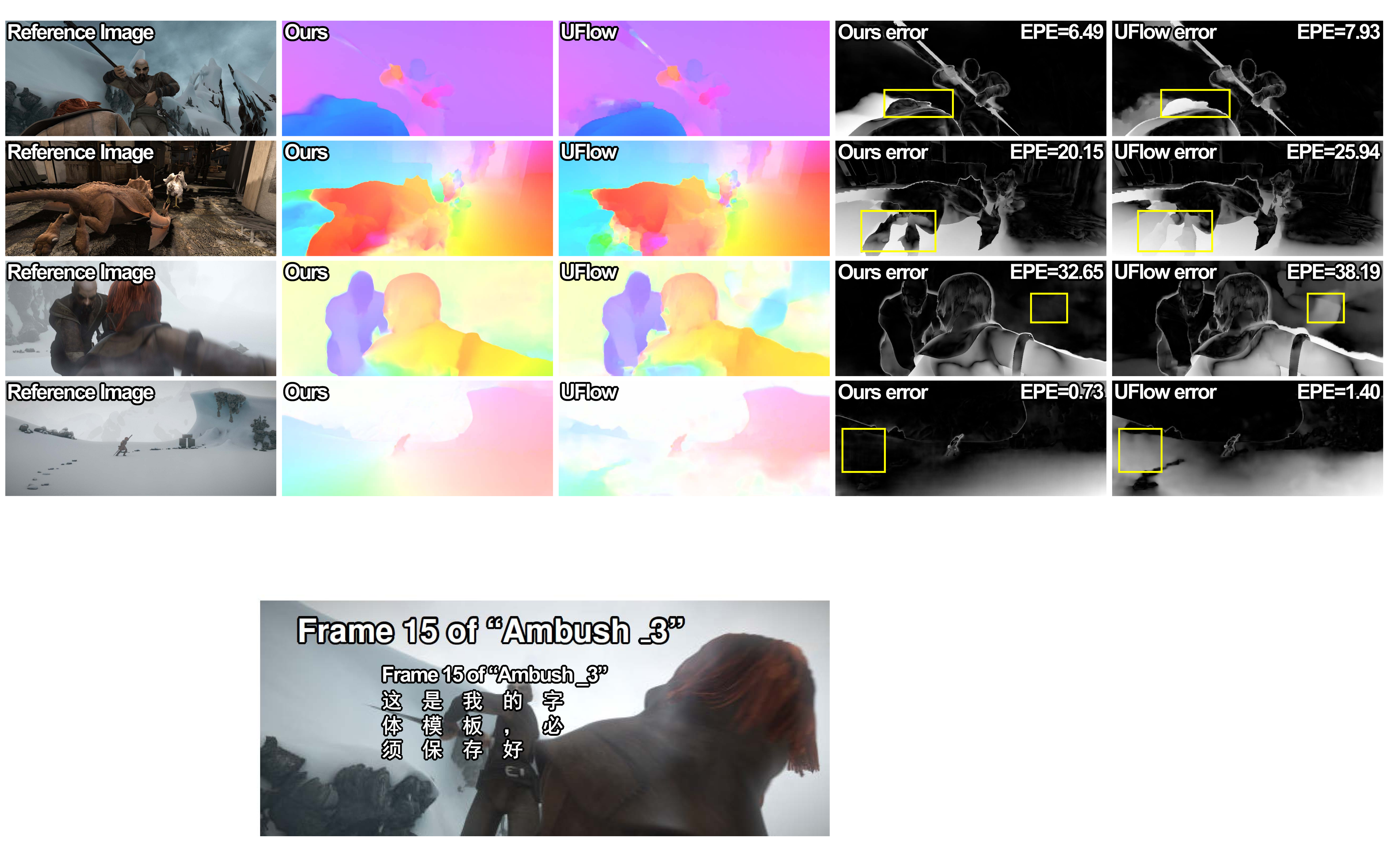}
		\label{fig:results_pk_benchmark_sintel}
	}
	\caption{Visual comparison of our method with the state-of-the-art method UFlow~\cite{jonschkowski2020matters} on KITTI (a) and Sintel (b) benchmarks. The error maps visualized by the benchmark websites are shown in the last two columns with obvious difference regions marked by yellow boxes. }
	\label{fig:results_pk_benchmark}
\end{figure*}

\subsection{Loss Guidance at Pyramid Levels}\label{sec:algo_loss_functions}
In our framework, we use several losses to train the pyramid network: the unsupervised optical flow losses for the final output flow and the pyramid distillation loss for the intermediate flows at different pyramid levels.

\subsubsection{Unsupervised Optical Flow Loss}\label{sec:algo_unsupervised_photometric_loss}
To learn the flow estimation model $\mathcal{H}$ in unsupervised setting, we use the photometric loss $\mathcal{L}_m$ based on the brightness constancy assumption that the same objects in $I_t$ and $I_{t+1}$ must have similar intensities. However, some regions may be occluded by moving objects, so that their corresponding regions do not exist in another image. 
Since the photometric loss can not work in these regions, we only add $\mathcal{L}_m$ on the non-occluded regions. The photometric loss can be formulated as follows:
\begin{align}
	\mathcal{L}_m=\frac{\sum_{\bm{p}}{\Psi\Big(I_t(\bm{p})-I_{t+1}\big(\bm{p}+V_{f}(\bm{p})\big)\Big) \cdot M_{t}(\bm{p})}}{\sum_{\bm{p}}{M_{1}(\bm{p})}}, \label{eq:photometric_loss_occ} 
\end{align}
where $M_{t}$ is the occlusion mask and $\Psi$ is the robust penalty function~\cite{Pengpeng2019}: $\Psi(x)=(|x|+\epsilon)^q$ with $q, \epsilon$ being 0.4 and 0.01. In the occlusion mask $M_{t}$, which is estimated by forward-backward checking~\cite{unflow_2018aaai}, $1$ represents the non-occluded pixels in $I_t$ and $0$ for those are occluded. 

To improve the performance, some previously effective unsupervised components are also added, including smooth loss~\cite{wang2018} $\mathcal{L}_{s}$, 
census loss~\cite{unflow_2018aaai} $\mathcal{L}_{c}$,
augmentation regularization loss~\cite{liu2020learning} $\mathcal{L}_{a}$, and boundary dilated warping loss~\cite{luo2020occinpflow} $\mathcal{L}_{b}$. 
For simplicity, we omit these components. Please refer to previous works for details. 
The capability of these components will be discussed in Sec.~\ref{sec:Ablation Study}. 



\begin{table*}[ht]
	\centering
	\resizebox*{0.94 \textwidth}{!}{
		\begin{tabular}{
				>{\centering\arraybackslash}p{0.3cm}
				p{3.2cm} 
				>{\centering\arraybackslash}p{0.8cm} 
				>{\centering\arraybackslash}p{0.8cm} 
				>{\centering\arraybackslash}p{0.8cm} 
				>{\centering\arraybackslash}p{2.2cm} 
				>{\centering\arraybackslash}p{0.8cm} 
				>{\centering\arraybackslash}p{0.8cm} 
				>{\centering\arraybackslash}p{0.8cm} 
				>{\centering\arraybackslash}p{0.8cm} 
			}
			\toprule
			\multicolumn{2}{c}{\multirow{2}{*}{Method}} & \multicolumn{2}{c}{KITTI 2012} & \multicolumn{2}{c}{KITTI 2015} & \multicolumn{2}{c}{Sintel Clean}&\multicolumn{2}{c}{Sintel Final}\\
			\cmidrule(lr){3-4} \cmidrule(lr){5-6} \cmidrule(lr){7-8} \cmidrule(lr){9-10}
			&&train & test &train & test (F1-all) &train &test &train &test
			\\
			\midrule
			\multirow{9}{*}{\rotatebox{90}{Supervised}}
			& FlowNetS~\cite{Flownet_flyingchairs}    &8.26  &  --  &  --  &   --  & 4.50 &7.42  &5.45  &8.43  \\
			& FlowNetS+ft~\cite{Flownet_flyingchairs} &7.52  &9.1   &  --  &   --  &(3.66)&6.96  &(4.44)&7.76  \\
			& SpyNet~\cite{spynet2017}                &9.12  &  --  &  --  &   --  &4.12  &6.69  &5.57  &8.43  \\
			& SpyNet+ft~\cite{spynet2017}             &8.25  &10.1  &  --  &35.07\%&(3.17)&6.64  &(4.32)&8.36  \\
			& LiteFlowNet~\cite{LiteFlowNet}          &4.25  &  --  &10.46 &   --  & 2.52 & -- & 4.05 &--  \\
			& LiteFlowNet+ft~\cite{LiteFlowNet}       &(1.26)& 1.7  &(2.16) &10.24\%&(1.64)& 4.86 &(2.23)&6.09  \\
			& PWC-Net~\cite{pwc_net}                  & 4.14 &  --  &10.35 &   --  & 2.55 &  --  & 3.93 &  --  \\
			& PWC-Net+ft~\cite{pwc_net}               &(1.45)& 1.7  &(2.16)&9.60\% &(1.70)& 3.86 &(2.21)&5.13  \\
			& IRR-PWC+ft~\cite{irrpwc}                &  --  &  --  &(1.63)&7.65\% &(1.92)& 3.84 &(2.51)&4.58  \\
			& RAFT~\cite{raft2020}                    &  --  &  --  & 5.54 &  --   & 1.63 &   -- & 2.83 & --  \\
			& RAFT-ft~\cite{raft2020}                 &  --  &  --  &  --  &6.30\% &  --  & 2.42 &  --  & 3.39  \\
			\midrule
			\multirow{12}{*}{\rotatebox{90}{Unsupervised}}
			& BackToBasic~\cite{Jason2016}               & 11.30& 9.9 &  --  &   --  &  --  &  --  &  --  &  --  \\
			& DSTFlow~\cite{Ren2017aaai}                 & 10.43& 12.4& 16.79& 39\%  &(6.16)&10.41 &(6.81)&11.27 \\
			& UnFlow~\cite{unflow_2018aaai}              & 3.29 &  --  & 8.10 & 23.3\%&  --  &9.38  &(7.91)&10.22 \\
			& OAFlow~\cite{wang2018}                     & 3.55 & 4.2 & 8.88 & 31.2\%&(4.03)&7.95  &(5.95)&9.15  \\
			& Back2Future~\cite{unflow_multi_occ}        &  --  &  --  & 6.59 &22.94\%&(3.89)&7.23  &(5.52)&8.81  \\
			& NLFlow~\cite{tip2020_nonlocalflow}         &3.02  &4.5   &6.05  &22.75\%&(2.58)&7.12  &(3.85)&8.51  \\
			& DDFlow~\cite{Pengpeng2019}                 &2.35  &3.0   &5.72  &14.29\%&(2.92)&6.18 &(3.98)&7.40  \\
			& EpiFlow~\cite{Epipolar_flow_2019cvpr}      &(2.51)&3.4   &(5.55)&16.95\%&(3.54)&7.00  &(4.99)&8.51  \\
			& SelFlow~\cite{Liu2019CVPR}                 &1.69  &2.2   &4.84  &14.19\%&(2.88)&6.56  &(3.87)&6.57  \\
			& STFlow~\cite{tip2020_nonlocalflow}         &1.64  &1.9   &3.56  &13.83\%&(2.91)&6.12  &(3.59)&6.63  \\
			& ARFlow~\cite{liu2020learning}              &\textcolor{blue}{1.44} &\textcolor{blue}{1.8} &2.85  &11.80\%&(2.79)&\textcolor{blue}{4.78}   &(3.87)&\textcolor{blue}{5.89}   \\
			& SimFlow~\cite{simFlow2020eccv}        & --  & --   &5.19  &13.38\%&(2.86)&5.92  &(3.57)&6.92  \\
			& UFlow~\cite{jonschkowski2020matters}       &1.68  &  1.9 &\textcolor{blue}{2.71}  &\textcolor{blue}{11.13\%}&\textcolor{blue}{ (2.50)}&5.21  &\textcolor{blue}{(3.39)} &6.50  \\
			& Ours                                       & \textcolor{red}{1.27} & \textcolor{red}{1.4} & \textcolor{red}{2.45} & \textcolor{red}{9.38\%}&\textcolor{red}{(2.33)}&\textcolor{red}{4.68}  &\textcolor{red}{(2.67)}&\textcolor{red}{5.32}  \\
			\bottomrule
		\end{tabular}
	}
	\caption{Comparison with previous methods. We use the average EPE error (the lower the better) as evaluation metric for all the datasets except on KITTI 2015 benchmark test, where the F1 measurement (the lower the better) is used. 
		Missing entries `$-$' indicates that the result is not reported in the compared paper, and $(\cdot)$ indicates that the testing images are used during unsupervised training. The best unsupervised results are marked in red and the second best are in blue. Note that, for results of the supervised methods, `+ft' means the model is trained on the target domain, otherwise, the model is trained on synthetic datasets such as Flying Chairs~\cite{Flownet_flyingchairs} and Flying Chairs occ~\cite{irrpwc}. For unsupervised methods, we report the performance of the model trained using images from target domain. 
	}
	\label{table:comparision_with_existing_method}
\end{table*}
\subsubsection{Pyramid Distillation Loss}\label{sec:algo_pyramid_distillation_loss}
To learn intermediate flow for each pyramid level, we propose to distillate the finest output flow to the intermediate ones by our pyramid distillation loss $\mathcal{L}_{d}$.
Intuitively, this is equivalent to calculating all the unsupervised losses on each of the intermediate outputs. However, the photometric consistency measurement is not accurate enough for optical flow learning at low resolutions~\cite{jonschkowski2020matters}. As a result, it is inappropriate to enforce unsupervised losses at intermediate levels, especially at the lower pyramid levels. Therefore, we propose to use the finest output flow as pseudo labels and add supervised losses instead of unsupervised losses for intermediate outputs. 

To calculate $\mathcal{L}_{d}$, we directly downsample the final output flow and evaluate its difference with the intermediate flows. Since occlusion regions are excluded from $\mathcal{L}_m$, flow estimation in occlusion regions is noisy. In order to eliminate the influence of these noise regions in the pseudo label, we also downsample the occlusion mask $M_t$ and exclude occlusion regions from $\mathcal{L}_{d}$. Thus, our pyramid distillation loss can be formulated as follow:
\begin{eqnarray}\label{eq:pyramid-distillation-loss}
	\mathcal{L}_{d} = \sum_{i=0}^{N} \sum_{\bm{p}}\Psi\big(V_{f}^{i}-\mathcal{S}_{\downarrow}(s_i,V_{f})\big) \cdot \mathcal{S}_{\downarrow}(s_i, M_{t}), 
\end{eqnarray}
where $s_i$ is the scale magnification of pyramid level $i$ and $\mathcal{S}_{\downarrow}$ is the downsampling function. 

Eventually, our training loss $\mathcal{L}$ is formulated as follows: 
\begin{eqnarray}\label{eq:total-training-loss}
	\mathcal{L}=\mathcal{L}_m+\lambda_d\mathcal{L}_d + \lambda_s\mathcal{L}_s + \lambda_c\mathcal{L}_c + \lambda_a\mathcal{L}_a + \lambda_b\mathcal{L}_b, 
\end{eqnarray}
where $\lambda_d$, $\lambda_s$, $\lambda_c$, $\lambda_a$ and $\lambda_b$ are hyper-parameters and we set $\lambda_d=0.01$, $\lambda_s=0.05$, $\lambda_c=1$, $\lambda_a=0.5$ and $\lambda_b=1$. 

\section{Experimental Results}\label{sec:results}
\subsection{Dataset and Implementation Details}\label{sec:Datasets}
We conduct experiments on three datasets: MPI-Sintel~\cite{Butler2012}, KITTI 2012~\cite{KITTI_2012} and KITTI 2015~\cite{KITTI_2015}. We use the same dataset setting as previous unsupervised methods~\cite{liu2020learning,jonschkowski2020matters}. 
For MPI-Sintel dataset, which contains $1,041$ training image pairs rendered in two different passes (`Clean' and `Final'), we use all the training images from both `Clean' and `Final' to train our model. For KITTI 2012 and 2015 datasets, we pretrain our model using $28,058$ image pairs from the KITTI raw dataset and then finetune our model on the multi-view extension dataset. The flow ground-truth is only used for validation. 

We implement our method with PyTorch, and complete the training in $1000k$ iterations with batch size of $4$. The total number of parameters of our model is $3.49\text{M}$, in which the proposed self-guided upsample module has $0.14\text{M}$ trainable parameters. Moreover, the running time of our full model is 0.05s for a Sintel image pair with resolution $436\times 1024$.
The standard average endpoint error (EPE) and the percentage of erroneous pixels (F1) are used as the evaluation metric of optical flow estimation.

\begin{table*}[ht]
	\centering
	\resizebox*{0.98 \textwidth}{!}{
		\begin{tabular}{
				>{\centering\arraybackslash}p{0.8cm} 
				>{\centering\arraybackslash}p{0.8cm} 
				>{\centering\arraybackslash}p{0.8cm} 
				>{\centering\arraybackslash}p{0.8cm} 
				>{\centering\arraybackslash}p{0.8cm} 
				>{\centering\arraybackslash}p{0.8cm} 
				>{\centering\arraybackslash}p{0.8cm} 
				>{\centering\arraybackslash}p{0.8cm} 
				>{\centering\arraybackslash}p{0.8cm} 
				>{\centering\arraybackslash}p{0.8cm} 
				>{\centering\arraybackslash}p{0.8cm} 
				>{\centering\arraybackslash}p{0.8cm} 
				>{\centering\arraybackslash}p{0.8cm} 
				>{\centering\arraybackslash}p{0.8cm} 
				>{\centering\arraybackslash}p{0.8cm} 
				>{\centering\arraybackslash}p{0.8cm} 
				>{\centering\arraybackslash}p{0.8cm} 
			}
			\toprule
			\\
			\multirow{2}{*}{CL} & \multirow{2}{*}{BDWL} & \multirow{2}{*}{ARL} & \multirow{2}{*}{SGU} & \multirow{2}{*}{PDL} & \multicolumn{3}{c}{KITTI 2012} & \multicolumn{3}{c}{KITTI 2015} & \multicolumn{3}{c}{Sintel Clean} & \multicolumn{3}{c}{Sintel Final} \\
			\cmidrule(lr){6-8} \cmidrule(lr){9-11} \cmidrule(lr){12-14} \cmidrule(lr){15-17}
			&                      &                      &                      &                      & ALL      & NOC         & OCC         & ALL         &    NOC      &    OCC      &    ALL       &    NOC       &    OCC      &    ALL       &    NOC       &    OCC      \\
			\midrule
			&                      &                      &                      &                      &   4.52   &    1.76     &   19.63     &   7.58      &   2.46      &  30.43      &   (3.52)       &   (1.87)       &  (12.93)      &   (4.19)       &   (2.59)       &  (13.64)      \\
			\Checkmark      &                      &                      &                      &                      &   3.39   &    1.09     &   16.58     &   6.89      &   2.20      &  28.12      &   (3.41)       &   (1.62)       &  (13.51)      &   (3.85)       &   (2.17)       &  (13.71)      \\
			\Checkmark      &     \Checkmark       &                      &                      &                      &   1.42   &    0.91     &   4.39      &   3.00      &   2.12      &   6.89      &   (2.84)       &   (1.50)       &  (10.63)      &   (3.60)       &   (2.28)       &  (11.52)      \\
			\Checkmark      &     \Checkmark       &     \Checkmark       &                      &                      &   1.37   &    0.93     &   3.98      &   2.64      &   1.96      &   6.01      &   (2.61)       &   (1.33)       &  (10.14)      &   (3.17)       &   (1.92)       &  (10.70)      \\
			\Checkmark      &     \Checkmark       &     \Checkmark       &     \Checkmark       &                      &   1.33   &    0.88     &   4.00      &   2.56      &   1.91      &   5.35      &   (2.46)       &   (1.17)       &   (9.89)      &   (2.79)       &   (1.53)       &  (10.28)      \\
			\Checkmark      &     \Checkmark       &     \Checkmark       &                      &     \Checkmark       &   1.36   &    0.91     &   4.03      &   2.61      &   1.96      &   5.52      &   (2.53)       &   (1.23)       &  (10.12)      &   (2.93)       &   (1.67)       &  (10.38)      \\
			\Checkmark      &     \Checkmark       &     \Checkmark       &     \Checkmark       &     \Checkmark    &\textbf{1.27}&\textbf{0.85}&\textbf{3.77}&\textbf{2.45}&\textbf{1.87}&\textbf{5.32}&\textbf{(2.33)} &\textbf{(1.07)} &\textbf{(9.66)}&\textbf{(2.63)} &\textbf{(1.39)} &\textbf{(9.91)}\\
			\bottomrule
		\end{tabular}
	}
	\caption{Ablation study of the unsupervised components. CL: census loss~\cite{unflow_2018aaai}, BDWL: boundary dilated warping loss~\cite{luo2020occinpflow}, ARL: augmentation regularization loss~\cite{liu2020learning}, SGU: self-guided upsampling, PDL: pyramid distillation loss. The best results are marked in bold.
	}
	\label{table:abl_components}
\end{table*}

\subsection{Comparison with Existing Methods}\label{sec:Comparison with Existing Methods}
We compare our method with existing supervised and unsupervised methods on leading optical flow benchmarks. Quantitative results are shown in Table~\ref{table:comparision_with_existing_method}, where our method outperforms all the previous unsupervised methods on all the datasets. In Table~\ref{table:comparision_with_existing_method}, we mark the best results by red and the second best by blue in unsupervised methods. 
\vspace{-10pt} 

\paragraph{Comparison with Unsupervised Methods.}
On KITTI 2012 online evaluation, our method achieves EPE=$1.4$, which improves the EPE=$1.8$ of the previous best method ARFlow~\cite{liu2020learning} by $22.2\%$. 
Moreover, on KITTI 2015 online evaluation, our method reduces the F1-all value of $11.13\%$ in UFlow~\cite{jonschkowski2020matters} to $9.38\%$ with $15.7\%$ improvement. On the test benchmark of 
MPI-Sintel dataset, we achieve EPE=$4.68$ on the `Clean' pass and EPE=$5.32$ on the `Final' pass, both outperforming all the previous methods. 
Some qualitative comparison results are shown in Fig.~\ref{fig:results_pk_benchmark}, where our method produces more accurate results than the state-of-the-art method UFlow~\cite{jonschkowski2020matters}. 
\vspace{-10pt} 

\paragraph{Comparison with Supervised Methods. }
As shown in Table~\ref{table:comparision_with_existing_method}, representative supervised methods are also reported for comparison. 
In practical applications where flow ground-truth is not available, the supervised methods can only train models using synthetic datasets. In contrast, unsupervised methods can be directly implemented using images from the target domain. 
As a result, on KITTI and Sintel Final datasets, our method outperforms all the supervised methods trained on synthetic datasets, especially in real scenarios such as the KITTI 2015 dataset. 

As for the in-domain ability, our method is also comparable with supervised methods. Interestingly, on KITTI 2012 and 2015 datasets, our method achieve EPE=$1.4$ and F1=$9.38\%$, which outperforms classical supervised methods such as PWC-Net~\cite{pwc_net} and LiteFlowNet~\cite{LiteFlowNet}.

\begin{figure*}
	\centering
	\includegraphics[width=0.98\textwidth]{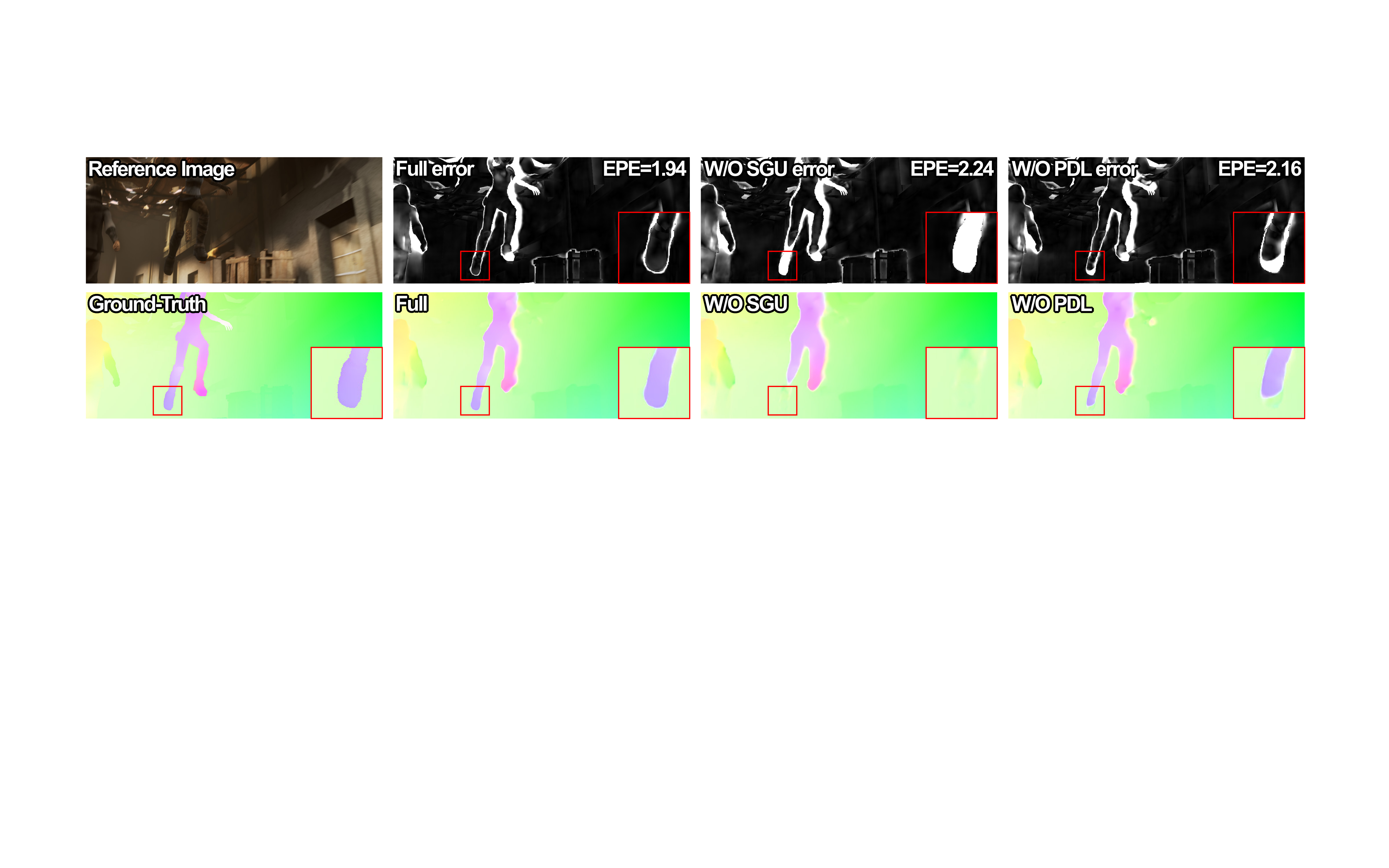}
	\caption{Visual results of removing the SGU or PDL from our full method on Sintel dataset. The room in flows and error maps are shown in the right corner of each sample.  
	}\label{fig:ablation_sgu_and_pdl}
\end{figure*}

\subsection{Ablation Study}\label{sec:Ablation Study}
To analyze the capability and design of each individual component, we conduct extensive ablation experiments on the train set of KITTI and MPI-Sintel datasets following the setting in\cite{Liu2019CVPR,simFlow2020eccv}. The EPE error over all pixels (ALL), non-occluded pixels (NOC) and occluded pixels (OCC) are reported for quantitative comparisons.
\vspace{-10pt} 

\begin{table}[t]
	\centering
	\resizebox*{1.0\linewidth}{!}{
		\begin{tabular}{
				p{1.4cm} 
				>{\centering\arraybackslash}p{1.8cm} 
				>{\centering\arraybackslash}p{1.8cm} 
				>{\centering\arraybackslash}p{1.8cm} 
				>{\centering\arraybackslash}p{1.8cm} 
			}
			\toprule
			Method              &  KITTI 2012  &   KITTI 2015 & Sintel Clean  & Sintel Final  \\
			\midrule
			Bilinear                             &     1.36     &     2.61     &   (2.53)        &   (2.93)       \\
			JBU~\cite{kopf2007JBU}               &     1.51     &     3.00     &   (2.66)        &   (2.98)      \\
			GF~\cite{he2010guided}               &     1.40     &     2.90     &   (2.72)        &   (2.92)       \\
			DJF~\cite{li2016deep}                &     1.36     &     2.79     &   (2.75)        &   (3.20)       \\
			DGF~\cite{wu2018fast}                &     1.41     &     3.14     &   (2.69)        &   (3.05)       \\
			PAC~\cite{su2019pac}                 &     1.42     &     2.65     &    (2.58)       &   (2.95)      \\
			SGU-FM                               &     1.35     &     2.60     &    (2.52)       &   (2.91)       \\
			SGU-M                                &     1.33     &     2.59     &    (2.41)       &   (2.86)       \\
			SGU                              & \textbf{1.27}& \textbf{2.45}& \textbf{(2.33)} & \textbf{(2.63)} \\
			\bottomrule
	\end{tabular}}
	\caption{Comparison of our SGU with different upsampling methods: the basic bilinear upsampling, image guided  upsampling methods including JBU~\cite{kopf2007JBU}, GF~\cite{he2010guided}, DJF~\cite{li2016deep}, DGF~\cite{wu2018fast} and PAC~\cite{su2019pac}, and the variants of SGU such as SGU-FM, where the interpolation flow and weight map are both removed, and SGU-M, where the only the interpolation map is removed. 
	}
	\label{table:ablation_study_with_upsample_structure}
\end{table}

\paragraph{Unsupervised Components. }
Several unsupervised components are used in our framework including census loss~\cite{unflow_2018aaai} (CL), boundary dilated warping loss~\cite{luo2020occinpflow} (BDWL), augmentation regularization loss~\cite{liu2020learning} (ARL), our proposed self-guided upsampling (SGU) and pyramid distillation loss (PDL). We assess the effect of these components in Table.~\ref{table:abl_components}. 
In the first row of Table.~\ref{table:abl_components}, we only use photometric loss and smooth loss to train the pyramid network with our SGU disabled. 
Comparing the first four rows in Table.~\ref{table:abl_components}, we can see that by combining CL, BDWL and ARL, the performance of optical flow estimation can be improved, which is equivalent to the current best performance reported in UFlow~\cite{jonschkowski2020matters}. Comparing the last four rows in Table.~\ref{table:abl_components}, we can see that: (1) the EPE error can be reduced by using our SGU to solve the bottom-up interpolation problem; (2) the top-down supervision information by our PDL can also improve the performance; (3) the performance can be further improved by combining the SGU and PDL. 

Some qualitative comparison results are shown in Fig.~\ref{fig:ablation_sgu_and_pdl}, where `Full' represents our full method, `W/O SGU' means the SGU module of our network is disabled and `W/O PDL' means the PDL is not considered during training. Comparing with our full method, the boundary of the predicted flow becomes blurry when SGU is removed while the error increases when PDL is removed. 
\vspace{-10pt}

\begin{table*}[ht]
	\centering
	\resizebox*{0.98\linewidth}{!}{
		\begin{tabular}{
				p{2.4cm} 
				>{\centering\arraybackslash}p{0.8cm} 
				>{\centering\arraybackslash}p{0.8cm} 
				>{\centering\arraybackslash}p{0.8cm} 
				>{\centering\arraybackslash}p{0.8cm} 
				>{\centering\arraybackslash}p{0.8cm} 
				>{\centering\arraybackslash}p{0.8cm} 
				>{\centering\arraybackslash}p{0.8cm} 
				>{\centering\arraybackslash}p{0.8cm} 
				>{\centering\arraybackslash}p{0.8cm} 
				>{\centering\arraybackslash}p{0.8cm} 
				>{\centering\arraybackslash}p{0.8cm} 
				>{\centering\arraybackslash}p{0.8cm} 
			}
			\toprule
			\multirow{2}{*}{Method}              &  \multicolumn{6}{c}{Sintel Clean train}  & \multicolumn{6}{c}{Sintel Final train}   \\
			\cmidrule(lr){2-7} \cmidrule(lr){8-13}
			&  $\times1$    &  $\times4$   &  $\times8$   &  $\times16$  &  $\times32$  &  $\times64$  &    $\times1$    &   $\times4$  &   $\times8$  &  $\times16$  &  $\times32$  &  $\times64$  \\
			\midrule
			w/o PL              &      (2.46)   &    (2.53)    &    (2.78)    &    (3.38)    &    (4.70)    &    (7.39)        &      (2.79)     &    (2.89)    &    (3.11)    &    (3.73)    &    (5.07)    &    (7.59)    \\
			PUL-up              &      (2.45)   &    (2.52)    &    (2.75)    &    (3.35)    &    (4.61)    &    (7.32)        &      (2.77)     &    (2.86)    &    (3.09)    &    (3.68)    &    (5.00)    &    (7.52)    \\ 
			PUL-down            &      (2.49)   &    (2.56)    &    (2.82)    &    (3.43)    &    (4.84)    &    (7.69)        &      (2.80)     &    (2.89)    &    (3.12)    &    (3.74)    &    (5.18)    &    (8.26)    \\ 
			PDL w/o occ         &      (2.37)   &    (2.42)    &    (2.61)    &    (3.15)    &    (4.17)    &    (6.31)        &      (2.73)     &    (2.81)    &    (3.00)    &    (3.59)    &    (4.82)    &    (7.16)    \\ 
			PDL                 &\textbf{(2.33)}&\textbf{(2.37)}&\textbf{(2.56)}&\textbf{(3.03)}&\textbf{(3.88)}&\textbf{(5.58)}&\textbf{(2.63)}&\textbf{(2.69)}&\textbf{(2.87)}&\textbf{(3.38)}&\textbf{(4.43)}&\textbf{(6.39)} \\
			\bottomrule
	\end{tabular}}
	\caption{Comparison of different pyramid losses: no pyramid loss (w/o PL), pyramid unsupervised loss by upsampling intermediate flows to image resolution to compute unsupervised objective functions (PUL-up) and by downsampling images to the intermediate resolution (PUL-down), our pyramid distillation loss without masking out occlusion regions (PDL w/o occ) and our pyramid distillation loss (PDL). All the intermediate output flows are evaluated on the train set of Sintel Clean and Final.
	}
	\label{table:ablation_study_with_pyramid_distillation_loss}
\end{table*}

\paragraph{Self-guided Upsample Module. }
There is a set of methods that use image information to guide the upsampling process, e.g.,
JBU~\cite{kopf2007JBU}, GF~\cite{he2010guided}, DJF~\cite{li2016deep}, DGF~\cite{wu2018fast} and PAC~\cite{su2019pac}. We implement them into our pyramid network and train with the same loss function for comparisons. The average EPE errors of the validation sets are reported in Table.~\ref{table:ablation_study_with_upsample_structure}. As a result, our SGU is superior to the image guided upsampling methods. 
The reason lies in two folds: (1) the guidance information that directly extracted from images may not be favorable to the unsupervised learning of optical flow especially for the error-prone occlusion regions; (2) our SGU can capture detail matching information by learning from the alignment features which are used to compute optical flow by the decoder. 


In the last three rows of Table.~\ref{table:ablation_study_with_upsample_structure}, we also compare our SGU with its variants: (1) SGU-FM, where the interpolation flow and interpolation map are both removed so that the upsampled flow is directly produced by the dense block without warping and fusion in Fig.~\ref{fig:algo_our_upsample_model}; (2) SGU-M, where the interpolation map is disabled. 
Although the performance of SGU-FM is slightly better than the baseline bilinear method, it is poor than that of SGU-M, which demonstrates that using an interpolation flow to solve the interpolation blur is more effective than directly learning to predict a new optical flow. Moreover, the performance reduced as the interpolation map is removed from SGU, which demonstrates the effectiveness of the interpolation map. 
\vspace{-10pt}

\paragraph{Pyramid Distillation Loss. }
We compare our PDL with different pyramid losses in Table.~\ref{table:ablation_study_with_pyramid_distillation_loss}, where `w/o PL' means no pyramid loss is calculated, `PUL-up' and `PUL-down' represent the pyramid unsupervised loss by upsampling intermediate flows to the image resolution and by downsamling the images to the intermediate resolutions accordingly. `PDL w/o occ' means the occlusion masks on pyramid levels are disabled in our PDL. 
In `PUL-up' and `PUL-down', the photometric loss, smooth loss, census loss and boundary dilated warping loss are used for each pyramid level and their weights are tuned to our best in the experiments. 
To eliminate variables, the occlusion masks used in `PUL-up' and `PUL-down' are calculated by the same method as in our PDL. 
As a result, model trained by our pyramid distillation loss can generate better results on each pyramid level than by pyramid unsupervised losses. 
This is because our pseudo labels can provide better supervision on low resolutions than unsupervised losses. 
Moreover, the error increased when the occlusion mask is disabled in our PDL, indicating that excluding the noisy occlusion regions can improve the quality of the pseudo labels.

\section{Conclusion}\label{sec:conclu}
We have proposed a novel framework for unsupervised learning of optical flow estimation by bottom-up and top-down optimize of the pyramid levels. 
For the interpolation problem in the bottom-up upsampling process of pyramid network, we proposed a self-guided upsample module to change the interpolation mechanism. 
For the top-down guidance of the pyramid network, we proposed a pyramid distillation loss to improve the optical flow learning on intermediate levels of the network.  
Extensive experiments have shown that our method can produce high-quality optical flow results, which outperform all the previous unsupervised methods on multiple leading benchmarks. 
\vspace{-14pt}

\paragraph{Acknowledgement:} This research was supported by National Natural Science Foundation of China (NSFC) under grants No.61872067 and No.61720106004.

%

\end{document}